\def\year{2022}\relax
\documentclass[letterpaper]{article} 
\usepackage{aaai22}  
\usepackage{times}  
\usepackage{helvet}  
\usepackage{courier}  
\usepackage[hyphens]{url}  
\usepackage{graphicx} 
\usepackage{natbib}  
\usepackage{caption} 
\DeclareCaptionStyle{ruled}{labelfont=normalfont,labelsep=colon,strut=off} 
\usepackage{algorithm}
\usepackage{algorithmic}
\usepackage{threeparttable}
\usepackage{tabularx}
\usepackage{newfloat}
\usepackage{listings}
\usepackage{amsfonts}
\floatstyle{ruled}
\newfloat{listing}{tb}{lst}{}
\floatname{listing}{Listing}
\title{Handwritten Mathematical Expression Recognition via Attention Aggregation based Bi-directional Mutual Learning}
\author {
    Xiaohang Bian\textsuperscript{\rm 1},
    Bo Qin \textsuperscript{\rm 2},
    Xiaozhe Xin \textsuperscript{\rm 2},
    Jianwu Li \textsuperscript{\rm 1}\footnote{Corresponding author: \textit{Jianwu Li}.},
    Xuefeng Su \textsuperscript{\rm 2},
    Yanfeng Wang\textsuperscript{\rm 2}, 
}
\affiliations {
    \textsuperscript{\rm 1} 
School of Computer Science and Technology, Beijing Institute of Technology, China\\
    \textsuperscript{\rm 2} AI Interaction Department, Tencent, China\\
    \{xhang.bian,ljw\}@bit.edu.cn, 
    \{bqin,georgexin,felixsu,shawndwang\}@tencent.com, 
}

\usepackage{bibentry}

\usepackage{multirow}
\usepackage{tabularx}

\begin{document}

\maketitle
\begin{abstract}

Handwritten mathematical expression recognition aims to automatically generate LaTeX sequences from given images. Currently, attention-based encoder-decoder models are widely used in this task. They typically generate target sequences in a left-to-right (L2R) manner, leaving the right-to-left (R2L) contexts unexploited. In this paper, we propose an Attention aggregation based Bi-directional Mutual learning Network (ABM) which consists of one shared encoder and two parallel inverse decoders (L2R and R2L). The two decoders are enhanced via mutual distillation, which involves one-to-one knowledge transfer at each training step, making full use of the complementary information from two inverse directions. Moreover, in order to deal with mathematical symbols in diverse scales, an Attention Aggregation Module (AAM) is proposed to effectively integrate multi-scale coverage attentions.
Notably, in the inference phase, given that the model already learns knowledge from two inverse directions, we only use the L2R branch for inference, keeping the original parameter size and inference speed.
Extensive experiments demonstrate that our proposed approach achieves the recognition accuracy of 56.85 $\%$ on CROHME 2014, 52.92 $\%$ on CROHME 2016, and 53.96 $\%$ on CROHME 2019 without data augmentation and model ensembling, substantially outperforming the state-of-the-art methods. The source code is available in https://github.com/XH-B/ABM. 


\end{abstract}

\section{Introduction}

Handwritten Mathematical Expression Recognition (HMER) has multiple application scenarios, such as intelligent education, human-computer interaction and academic paper writing auxiliary tools. Traditional methods generating LaTeX sequences from input images always depend on specially designed grammars~\cite{lavirotte1998mathematical,chan2001error,maclean2013new}. 
These grammars need extensive prior knowledge to define mathematical expression structures, symbols' position relationship, and corresponding parsing algorithms, so that they cannot recognize complex mathematical expressions.

Recently, attention based encoder-decoder models have been applied to HMER, due to its excellence in machine translation~\cite{cho2014learning}, speech recognition~\cite{bahdanau2016end}, segmentation~\cite{zhou2020matnet,zhou2021cascaded,wang2021hierarchical}. These attention-based methods \cite{deng2017image,le2017training,zhang2017watch,zhang2018deep,zhang2018multi,wu2018image,le2019pattern,li2020improving,wu2020handwritten} are remarkably superior to grammar-based ones. 
\begin{figure*}[!t]
    \centering
    \includegraphics[width=\textwidth]{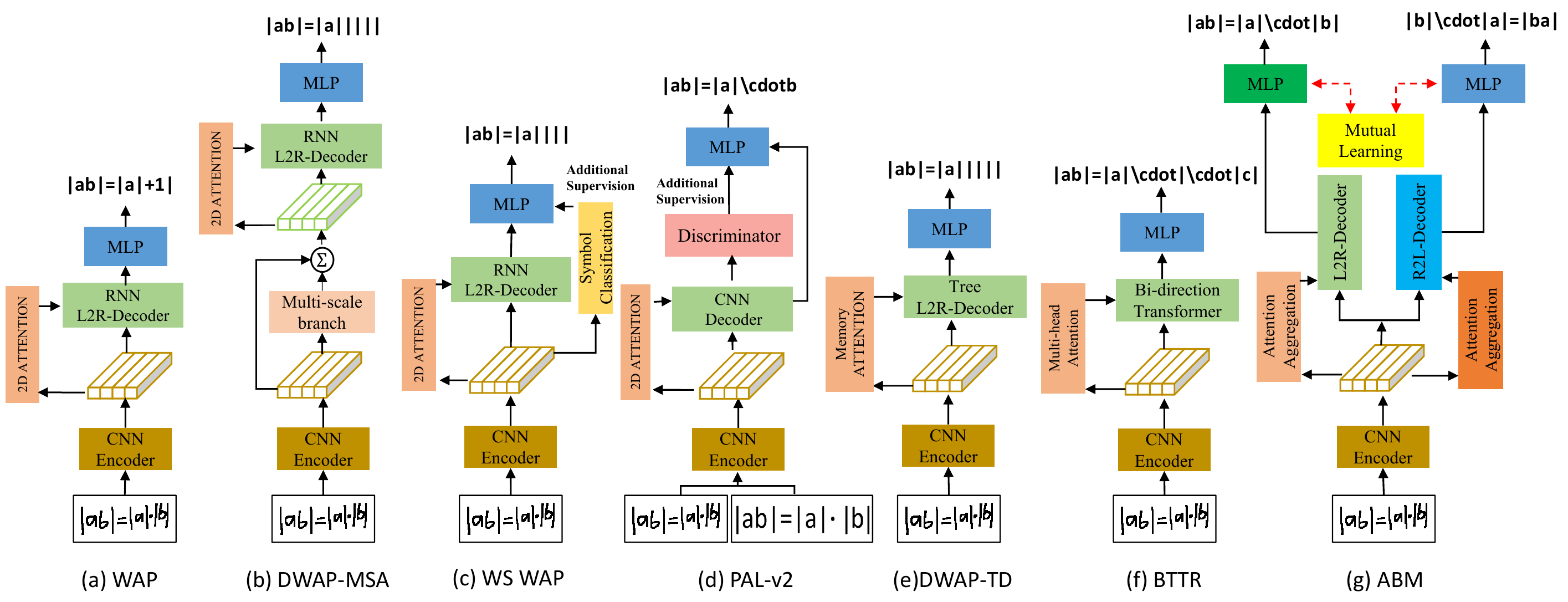}
    \caption{Typical architectures and our proposed model for HMER. (a) is 2D attention based encoder-decoder framework for HMER~\cite{zhang2017watch}. (b) uses multi-scale features~\cite{zhang2018deep}. (c) introduces a symbol classification network~\cite{truong2020improvement}. (d) utilizes existing printed expressions~\cite{wu2020handwritten}. (e) introduces a tree decoder~\cite{zhang2020tree}. (f) uses a transformer decoder~\cite{zhao2021handwritten}. (g) is our proposed model with two inverse decoders learning from each other to enhance their decoding ability.}
    \label{fig:compare_network}
\end{figure*}
For example, WAP~\cite{zhang2017watch} first introduces the 2D coverage attention to solve the problem of lacking coverage, as shown in Fig.~\ref{fig:compare_network} (a). The coverage attention is the sum of past attentions aiming to keep track of past alignment information, such that the attention model can be guided to assign higher attention probabilities to the untranslated regions of images. Nevertheless, one main limitation of the coverage attention is that it only employs historical alignment information, disregarding future information (untranslated regions). For example, many mathematical expressions are of symmetrical structures, where the left “$\{$” and right “$\}$” braces always appear together or sometimes far apart. And some symbols in an equation are correlated, such as “$\int$” and “$dx$”. Most methods only use the left-to-right coverage attention to identify the current symbol, ignoring the fact that the future information from the right is also important, which may cause the problem of attention drift. And the captured dependence information between the current symbol and previous symbols becomes weaker as their distances increase. Therefore, they may not adequately exploit long-distance correlation or grammatical specification of mathematical expressions~\cite{zhao2021handwritten}. BTTR~\cite{zhao2021handwritten} uses a transformer decoder with two directions to handle attention drift (Fig.~\ref{fig:compare_network}(f)), but there is no explicitly supervised information for BTTR to learn from the reversed direction, and BTTR aligns attention without a coverage mechanism, making it still suffer from some limitations in recognizing long formulas.
Besides, variable scales of characters in a mathematical expression may result in recognition difficulty or uncertainty. DWAP-MSA~\cite{zhang2018deep} attempts to encode multi-scale features to alleviate this problem. However, they do not scale the local receptive field, but only scale the feature map, making it impossible to accurately attend to small characters during recognition.

Thereby, we propose a novel framework with \textbf{A}ttention aggregation and \textbf{B}i-directional \textbf{M}utual learning (ABM) for HMER, as shown in Fig.~\ref{fig:compare_network}(g). Specially, our framework includes three modules: Feature Extraction, Attention Aggregation and Bi-directional Mutual Learning.
(1) In Feature Extraction module, we use DenseNet as feature extractor as it has proved to be effective in WAP~\cite{zhang2017watch}. 
(2) In Attention Aggregation module, we propose multi-scale coverage attention to recognize characters of different sizes in mathematical expressions, thereby improving the recognition accuracy at the current moment and alleviating the problem of error accumulation.
(3) In Bi-directional Mutual Learning module, we propose a novel decoder framework with two parallel decoder branches in opposite decoding directions (L2R and R2L) and use mutual distillation to learn from each other. Specifically, this framework helps the current coverage attention to capitalize upon historical and future information sufficiently at the same time, so as to better determine the current attention position. Therefore, each branch can learn more complementary context information and explore long-distance dependency information through step-by-step mutual learning, leading to stronger decoding ability. Note that while we use two decoders for training, we only use one L2R branch for inference.
Our contributions are summarized as follows:

(1) We propose a novel bi-directional mutual learning framework with a shared encoder and two inverse decoders, to better learn complementary context information. To our best knowledge, we are the first to introduce mutual learning into HMER.

(2) We propose a 
multi-scale coverage attention mechanism to better recognize the symbols with variable scales in an expression.

(3) Comprehensive experiments show that ABM greatly surpasses the state-of-the-art methods on CROHME 2014, 2016 and 2019, respectively. And the ABM framework can be applied to various decoders including GRU, LSTM, and Transformer.

\section{Related work}
\subsection{Methods of HMER}
Traditional HMER methods require specially designed grammatical rules to represent the two-dimensional structural information of formulas, such as graph grammar~\cite{lavirotte1998mathematical}, attributive clause grammar~\cite{chan2001error}, relational grammar~\cite{maclean2013new} or probabilistic model based grammar~\cite{awal2014global,maclean2015bayesian,alvaro2016integrated}.

In recent years, with the success of sequence learning in various applications, such as machine translation~\cite{luong2014addressing} and speech recognition~\cite{bahdanau2016end}, the encoder-decoder framework has been widely used to solve image-to-sequence tasks. \cite{deng2017image} first introduced such a framework to HEMR, which uses Convolutional Neural Network (CNN) and Recurrent Neural Network (RNN)~\cite{kawakami2008supervised}
as an encoder for feature extraction, while the Gated Recurrent Unit (GRU)~\cite{chung2014empirical} is used as a decoder to recognize LaTeX characters.
Many approaches typically improve the encoder with stronger convolutional networks to strengthen feature extraction, such as introducing full convolutional networks~\cite{zhang2017watch,Wang2019MultimodalAN}, DenseNet~\cite{zhang2018multi,le2019pattern,truong2020improvement} and ResNet~\cite{li2020improving,Yan2021ConvMathAC}.
For the decoder, most methods design attention mechanisms to improve their translation. 
For example, the coverage attention mechanism \cite{zhang2017watch,zhang2018multi,li2020improving,le2020recognizing,truong2020improvement}, overcomes the under-parsing or over-parsing problem by considering past alignment probabilities.
Additionally, some methods improve their recognition performance by introducing additional data such as data augmentation, including the pattern generation strategy PGS~\cite{le2019pattern}, random scale enhancement SCDA~\cite{li2020improving}, as well as the use of printed expressions DLA~\cite{le2020recognizing}, PAL~\cite{wu2018image} and PAL-v2~\cite{wu2020handwritten}.
Besides, some methods also improve the decoder with transformer to utilize bi-directional information (BTTR)~\cite{zhao2021handwritten} or tree decoder to enhance the decoding ability to handle complex formulas (DWAP-TD)~\cite{zhang2020tree}. 
The structures of some models above are illustrated in Fig.~\ref{fig:compare_network}(a$\sim$f). 

\subsection{Mutual Learning}
Mutual learning refers to the process in which a group of models learn together during training. DML~\cite{zhang2018deep} first proposed the concept of mutual learning in the field of knowledge distillation, and improved the model generalization ability through collaborative training. Co-distillation~\cite{anil2018large} forces each network to maintain its diversity through distillation losses. ONE~\cite{lan2018knowledge} trains a multi-branch network and uses the predictions of these branches as soft goals to guide each branch network. CLNN~\cite{song2018collaborative} designs hierarchical multiple branches and uses corresponding zoom gradients. KDCL~\cite{guo2020online} proposes to integrate soft targets from multiple networks, and then supervise the learning of each network.
These methods have been widely tried on public classification datasets, and each mutual learning work is usually of single CNN-based architecture. 

However, our experimental results show that direct mutual learning between two encoder-decoder networks cannot achieve satisfactory recognition accuracy for the HMER task.
Therefore, we design a novel architecture consisting of a shared encoder and two inverse decoders that can be aligned in each decoding step and learn from each other to fully explore the specific features of mathematical expressions, such as symmetry and long-distance correlation. 




\section{Method}
\begin{figure*}[t]
    \centering
    \includegraphics[width=1\linewidth]{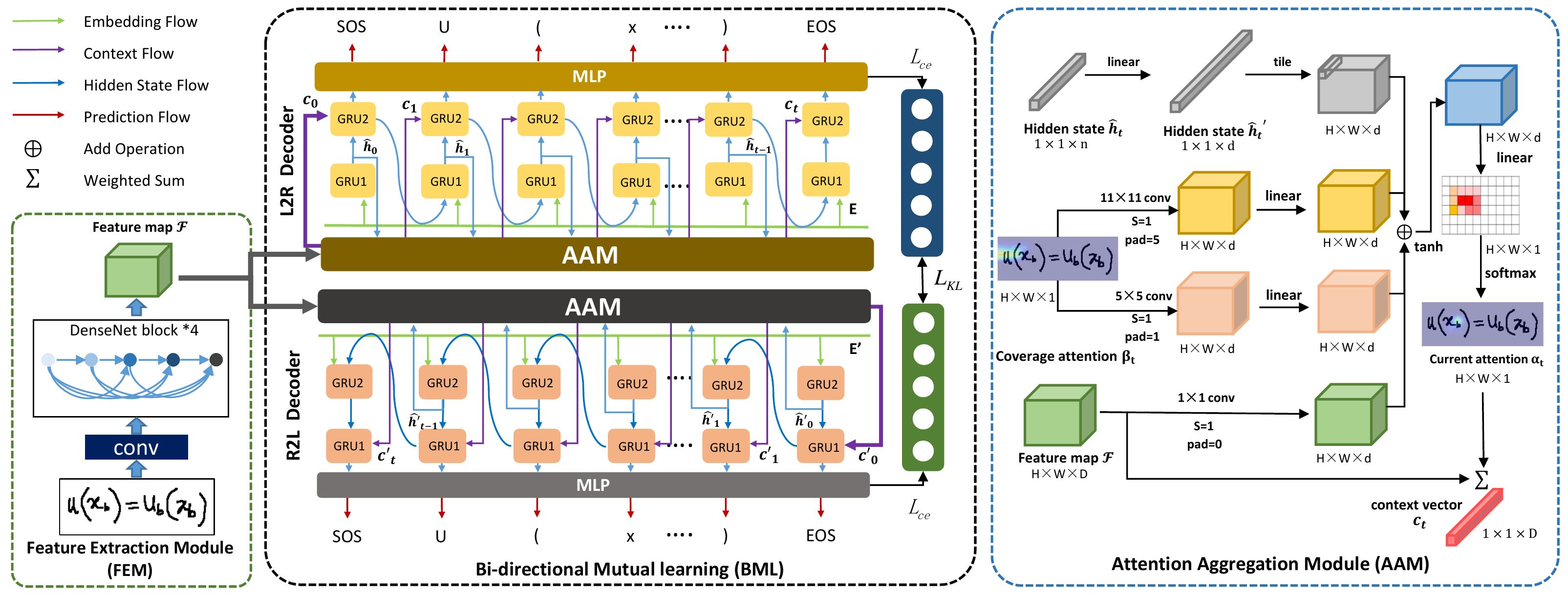}
    \caption{Architecture of our proposed model. An image is first input into the DenseNet to extract features, then two decoders separately generate the LaTeX sequences in two reverse directions. During decoding, two branches are trained by minimizing the distance between their predicted probabilities at each time step. An attention aggregation module is proposed to generate current attention by coverage attentions with different scales. The operation “tile” duplicates the hidden state vector $H\times W$ times. The “MLP” denotes the trainable multi-layer perception layers. The two branches do not share parameters during training.}
    \label{fig:network}
\end{figure*}

We propose a novel end-to-end architecture with Attention aggregation and Bi-directional Mutual learning (ABM) for HMER, as shown in Fig.~\ref{fig:network}. It mainly consists of three modules: 1) \textbf{F}eature \textbf{E}xtraction \textbf{M}odule (FEM) that extracts feature information from a mathematical expression image.
2) \textbf{A}ttention \textbf{A}ggregation \textbf{M}odule (AAM), which integrates multi-scale coverage attentions to align historical attention information, and effectively aggregates different scales of features from various sizes of symbols in the decoding phase.
3) \textbf{B}i-directional \textbf{M}utual \textbf{L}earning Module (BML) is comprised of two parallel decoders with opposite decoding directions (L2R and R2L) to complement information reciprocally. During training, each decoder branch can learn not only the ground-truth LaTeX sequence but also the prediction of the other branch, thereby enhancing the decoding ability.
\subsection{Feature Extraction Module}
We use the densely connected convolutional network (DenseNet~\cite{huang2017densely}) as the encoder to extract features from an input image, similar to \cite{zhang2018multi}. The output is a three-dimensional feature map $\mathcal{F}$ of $H\times W \times D$, where $H$, $W$ and $D$ respectively denote height, width and channel of the feature map. Specially, we consider the output features as content information $\textbf{a}$ of $M$ dimensions, forming a vector $\textbf{a} = \{a_1,a_2,...,a_M\}$, where $a_i \in \mathbb{R^D},M=H\times W$. 

\subsection{Attention Aggregation Module}

The attention mechanism prompts a decoder to focus on a specific area of input image. Specially, coverage-based attention can better track alignment information and guide a model to assign higher attention probabilities to untranslated regions~\cite{zhang2017watch}. Inspired by the Inception module~\cite{szegedy2015going}, we propose the Attention Aggregation Module (AAM) to aggregate different receptive fields on coverage attention. Compared with traditional attention, AAM not only pays attention to the detailed features of local areas, but also the global information on larger receptive fields. Therefore, AAM will generate finer information alignment and help the model capture more accurate spatial relationships. Note that our AAM is different from DWAP-MSA~\cite{zhang2018deep}, 
which proposes a dense encoder with multi-scale branches to generate low-resolution and high-resolution features, requiring more parameters and calculations. Fig.~\ref{fig:network} shows the details of our AAM that uses the hidden state $\hat{h}_t$, feature map $\mathcal{F}$ and coverage attention $\beta_t$ to compute the current attention weights $\alpha_t$ and then obtain context vector $c_t$:
\begin{equation}
    A_s  = U_s \beta_t,  A_l = U_l \beta_t,
\end{equation}
$U_s$ and $U_l$ denote the convolution operations of small and large kernel sizes ($e.g.$, 5, 11), respectively, and $\beta_t$ represents the sum of all past attention probabilities, which is initialized as a zero vector and then calculated by
\begin{equation}
\beta_t = \sum_{l=1}^{t-1}{\alpha_l},
\end{equation}
where $\alpha_l$ denotes the attention score at step $l$. The current attention map $\alpha_t$ is calculated by
\begin{equation}
\alpha_{t} = v_a^T \tanh(W_{\hat{h}} \hat{h_t}+ U_f \mathcal{F}+W_s A_s + W_l A_l),
\label{eqution:multi-scale}
\end{equation}
where $W_{\hat{h}} \in \mathbb{R}^{n \times d}$, $W_s\in \mathbb{R}^{1 \times d}$ and $W_l\in \mathbb{R}^{1 \times d}$ are trainable weight matrices, $U_f$ is 1$\times$1 convolution operation, and $\hat{h_t}$ denotes the hidden state generated from a GRU in Eq. (7).

The context vector is denoted as $c_t$ and is computed as a weighted sum of the feature content information $\textbf{a}$:
\begin{equation}
c_t = \sum_{i=1}^{M} \alpha_{t,i}\textbf{a}_i,
\end{equation}
where $\alpha_{t,i}$ is the weight of the $i$-th feature of $\mathcal{F}$ at step $t$.

\subsection{Bi-directional Mutual Learning Module}

Given an input mathematical expression image, traditional HMER methods decode it from left to right (L2R)~\cite{zhang2017watch,zhang2018multi}, without sufficiently considering the long-distance dependence.
Therefore, we propose to leverage dual-stream decoders to translate the input image to LaTex sequences in two opposite directions (L2R and R2L), and then learn the decoding information from each other.
The two branches have the same architecture, but are merely different in their decoding directions.

For bi-directional training, we add $\langle sos \rangle$ and $\langle eos \rangle$ respectively as the start and end symbol of LaTeX sequences. Specially, for the target LaTeX sequence of length $T$, $\mathcal{Y}=\{Y_1,Y_2,...,Y_T\}$, we denote it from left to right (L2R) as $\overrightarrow{\mathcal{Y}}_{l2r}=\{\langle sos \rangle,Y_1,Y_2,...,Y_T,\langle eos \rangle\}$ and from right to left (R2L) as $\overleftarrow{\mathcal{Y}}_{r2l}=\{\langle eos \rangle,Y_T, Y_{T-1},...,Y_1,\langle sos \rangle\}$. 
The probabilities of the predicted symbols at step $t$ for the L2R and R2L branch, are computed as follows:
\begin{equation}
    p(\overrightarrow{y}_t|\overrightarrow{y}_{t-1}) = W_o max(W_y E \overrightarrow{y}_{t-1} + W_h h_t + W_t c_t),
\end{equation}
\begin{equation}
    p(\overleftarrow{y}_t|\overleftarrow{y}_{t-1}) = W^{'}_o max(W^{'}_y E^{'} \overleftarrow{y}_{t-1} + W^{'}_h h_t^{'} + W^{'}_t c_t^{'}),
\end{equation}
where $h_{t}$, $\overrightarrow{y}_{t}$ denote the current state and previous prediction output at step $t$ in the L2R branch. The mark $*^{'}$ denotes the R2L branch. $W_o\in \mathbb{R}^{K \times d}$, $W_y\in \mathbb{R}^{d \times n}$, $W_h\in \mathbb{R}^{d \times n}$ and $W_t\in \mathbb{R}^{d \times D}$ are trainable weight matrices. Let $d$, $K$ and $n$ denote the attention dimension, the number of symbol classes and GRU dimension, respectively. $E$ is an embedding matrix. $max$ denotes the maxout activation function. The hidden representations $\{h_1,h_2, ...,h_t\}$ are produced by:
\begin{equation}
    \hat{h_t} = f_1(h_{t-1},E \overrightarrow{y}_{t-1}),
    \label{equ:ht}
\end{equation}
\begin{equation}
    h_t = f_2(\hat{h_t},c_{t}),
\end{equation}
where $f_1$ and $f_2$ denote two unidirectional GRU models similar to~\cite{zhang2017watch}.

We define the probability of the L2R branch as $\overrightarrow{\mathcal{P}}_{l2r} =\{\langle sos \rangle,\overrightarrow{y}_1,\overrightarrow{y}_2,...,\overrightarrow{y}_T,\langle eos \rangle \}$, and R2L branch as $\overleftarrow{\mathcal{P}}_{r2l} =\{\langle eos \rangle,\overleftarrow{y}_1,\overleftarrow{y}_{2},...,\overleftarrow{y}_T,\langle sos \rangle\}$, where $\overrightarrow{y}_i \in \mathbb{R}^K$ is the predicted probability of label symbols when the $i$-th step decoding is performed.
In order to apply mutual learning to the prediction distributions from two branches, we need to align the LaTeX sequences generated by the L2R and R2L decoders. Specifically, we discard the first and last predictions ($\langle eos \rangle$ and $\langle sos \rangle$) to obtain $\overrightarrow{\mathcal{P}}^{'}_{l2r}$ and $\overleftarrow{\mathcal{P}}^{'}_{r2l}$, and then reverse $\overleftarrow{\mathcal{P}}^{'}_{r2l}$ to obtain $\overleftarrow{\mathcal{P}}^{*}_{r2l}=\{\overleftarrow{y}_T,\overleftarrow{y}_{T-1}, ...,\overleftarrow{y}_1 \}$. At the same time, Kullback-Leibler (KL) loss is introduced to quantify the difference in prediction distribution between them.
During training, we use the soft probabilities generated by the model to provide more information, similar to~\cite{zhang2018multi}. Thus, for $k$ categories, the soft probability from L2R branch is defined as:
\begin{equation}
\sigma(\overrightarrow{Z}_{i,k},S) = \frac{exp({\overrightarrow{Z}_{i,k}/S})}{\sum_{j=1}^K exp({\overrightarrow{Z}_{i,j}/S})},
\end{equation}
where $S$ denotes the temperature parameter for generating soft labels. 
The logits of the $i$-th symbol of this sequence calculated by the decoder network are defined as $\overrightarrow{Z}_i =\{z_1, z_2,...,z_{K}\}$. Our objective is to minimize the distance between the two branch probability distributions. Thus, the KL distance between $\overrightarrow{\mathcal{P}}^{'}_{l2r}$ and $\overleftarrow{\mathcal{P}}^{*}_{r2l}$ is computed as follows:
\begin{equation}
    L_{KL} = S^2 \sum_{i=1}^{T} \sum_{j=1}^{K}  \sigma(\overrightarrow{Z}_{i,j},S)log \frac{\sigma(\overrightarrow{Z}_{i,j},S)}{\sigma(\overleftarrow{Z}_{T+1-i,j},S)},
\end{equation}
where $S^2$ ensures that the ground-truth and the probability distribution from the other branch can make comparable contributions to model training~\cite{hinton2015distilling}, and $\overrightarrow{Z}_{i,j}$ and $\overleftarrow{Z}_{T+1-i,j}$ denote the logits from L2R and R2L branch, respectively.
\subsection{Loss Function}
Specially, for the target LaTex sequence of length $T$, $\overrightarrow{\mathcal{Y}}_{l2r}=\{\langle sos \rangle,Y_1,Y_2,...,Y_T,\langle eos \rangle\}$, we denote the corresponding one-hot ground-truth label at the $i$-th time step as $Y_i = \{x_1, x_2, ... ,x_K\}$ with $x_i \in\{0,1\}$. 
The softmax probability of the $k$-th symbol is computed as:
\begin{equation}
\overrightarrow{y}_{i,k} = \frac{exp({\overrightarrow{Z}_{i,k}})}{\sum_{j=1}^K exp({\overrightarrow{Z}_{i,j}})}.
\end{equation}
For multi-class classification, the cross-entropy losses between the target label and softmax probability for two branches are defined as:
\begin{equation}
    L_{ce}^{l2r} = \sum_{i=1}^T \sum_{j=1}^K -Y_{i,j} log(\overrightarrow{y}_{i,j}).
\end{equation}

\begin{equation}
    L_{ce}^{r2l} = \sum_{i=1}^T \sum_{j=1}^K -Y_{i,j} log(\overleftarrow{y}_{T+1-i,j}).
\end{equation}





The overall loss function is as follows:
\begin{equation}
    L = L_{ce}^{l2r} + L_{ce}^{r2l}+ \lambda L_{KL}, 
\end{equation}
where $\lambda$ is a hyper-parameter to balance the recognition loss and KL divergence loss.

\begin{table}[!t]
    \centering
    \begin{threeparttable}
    \begin{tabular}{c c c c c}
    \hline
    Dataset&methods&ExpRate & $\leq$1 error & $\leq$2 error\\
    \hline
     \multirow{11}*{2014} 
     &PAL & 39.66 & 56.80 & 685.11  \\
     &WAP&46.55&61.16&65.21 \\
     &PGS & 48.78 &66.13&73.94 \\
     & PAL-v2 & 48.88 & 64.50 & 69.78 \\
     & DWAP-TD &49.10&64.20&67.8 \\
     & DLA & 49.85& -&- \\
     & DWAP&50.60& 68.05   & 71.56   \\
     & DWAP-MSA& 52.80& 68.10 & 72.00 \\
     &WS WAP & 53.65 & -& - \\
     & BTTR & 53.96 & 66.02 & 70.28 \\
     \cline{2-5}
     &\textbf{ ABM}  & \textbf{56.85} & \textbf{73.73} & \textbf{81.24} \\
     
     \hline
     \multirow{10}*{ 2016} 
     & PGS & 36.27 & - & -\\
     & TOKYO & 43.94 & 50.91 & 53.70\\
     &WAP&44.55&57.10&61.55 \\
     & DWAP-TD & 48.50& 62.30 & 65.30 \\
     & DLA & 47.34& -&-\\
     & DWAP&47.43 &60.21 & 63.35 \\
     &PAL-v2 & 49.61 & 64.08 & 70.27 \\
     & DWAP-MSA& 50.10& 63.80 & 67.40\\
     &WS WAP & 51.96 & 64.34& 70.10 \\
     & BTTR & 52.31 & 63.90 & 68.61 \\
     \cline{2-5}
     & \textbf{ABM} & 52.92 & \textbf{69.66}& \textbf{78.73}\\
     \hline
     \multirow{6}*{ 2019} 
     &DWAP&47.70&59.50&63.30 \\
     & DWAP-TD &51.40&66.10&69.10 \\
     & BTTR & 52.96 & 65.97 & 69.14 \\
     \cline{2-5}
     &\textbf{ABM} & \textbf{53.96} &\textbf{71.06} &\textbf{ 78.65} \\
    \hline
    \end{tabular}
    \caption{Comparison with prior works (in $\%$). Note that our results are from L2R branch. The results shown in the upper are partly cited from their corresponding papers. }
    \label{tab:compare with prior work}
    \end{threeparttable}
\end{table}





\section{Experiments}
\subsection{Datasets and Metrics}
We train our models based on the CROHME 2014 competition dataset with 111 classes of mathematical symbols and 8836 handwritten mathematical expressions, and test our model on three public test datasets: CROHME 2014, 2016, and 2019, with 986, 1147 and 1199 expressions, respectively. 
We use two indicators to evaluate the models: (1) Expression level: ExpRate ($\%$), $\leq$ 1 error ($\%$), and $\leq$ 2 error ($\%$) represent expression recognition accuracy when zero to two structural or symbol errors can be tolerated. (2) Word level: Word Error Rate (WER($\%$))~\cite{klakow2002testing} is used to evaluate such errors as substitutions, deletions, and insertions in word level.

\subsection{Implementation Details}

\textbf{Setup:} Two different decoder branches in our model are set to different weight initialization
methods~\cite{glorot2010understanding,he2015delving}. 
For the decoder, we set $n$ = 256, $d$=512, $D$=684 and $K$=113 (adding $\langle sos \rangle$ and $\langle eos \rangle$ on 111 labels). In the loss function, $\lambda$ is set to 0.5.
\textbf{Training:} Our proposed method is optimized with Adadelta optimizer, and its learning rate starts from 1, decaying two times smaller when the WER does not decrease within 15 epochs. And the training will stop early when the learning rate drops 10 times. We set the batch size as 16. 
\textbf{Testing Platform:} All the models are trained/tested on a single NVIDIA V100 16GB GPU.

\begin{table}[t]
    \centering
    \begin{threeparttable}

    \begin{tabular}{l c c c c c}
    \hline
    Methods&ExpRate&$\leq$1 error & $\leq$2 error & WER \\
    \hline
      Baseline &50.60 &  68.05 & 71.56  &  13.12   \\
    +AAM & 52.64 & 68.62 &77.25 & 12.12 \\
    +BML & 55.23 & 72.58 & 79.08 & 10.25 \\
    ABM & \textbf{56.85} & \textbf{73.73} & \textbf{81.24} & \textbf{10.01} \\
    
    \hline
    \end{tabular}
    \begin{tablenotes}
     \item[1] “+” means to append to baseline model.
   \end{tablenotes}
    \caption{Ablation study (in $\%$). We evaluate AAM and BML modules on CROHME 2014 test dataset. }
    \label{tab:ablation study}
    \end{threeparttable}
\end{table}

\subsection{Comparison with Prior Works}
We compare the ABM with the previous state-of-the-arts, including PAL~\cite{wu2018image}, PAL-v2~\cite{wu2020handwritten}, WAP~\cite{zhang2017watch}, PGS~\cite{le2019pattern}, DWAP (WAP with DenseNet as encoder), DWAP-MSA (DWAP with multi-scale attention)~\cite{zhang2018deep}, DWAP-TD (DWAP with tree decoder)~\cite{zhang2020tree}, DLA~\cite{le2020recognizing}, WS WAP (weakly supervised WAP)~\cite{truong2020improvement} and BTTR ( bidirectionally trained transformer)~\cite{zhao2021handwritten}. To ensure the fairness of the performance comparison, all the methods we show do not use data augmentation. 
From Table~\ref{tab:compare with prior work}, we can observe that:
\textbf{(1)} The proposed model ABM significantly improves the recognition accuracy (ExpRate) and outperforms the baseline (DWAP) by 6.25$\%$, 5.49$\%$ and  6.26$\%$ on three test datasets, demonstrating the success of bi-directional mutual learning module and attention aggregation module in enhancing the prediction capacity.
\textbf{(2)} Compared with other methods, our ABM model is superior to the previous state-of-the-arts in terms of almost all metrics. BTTR uses a traditional transformer as a decoder, which can reduce the decoding errors of long sequences to a certain extent. However, the results of $\leq$ 1 error and $\leq$ 2 error show that the word error rate cannot be significantly reduced.
On CROHME 2014, our model is more accurate than BTTR by large margins of 2.89$\%$, 7.71$\%$ and 10.96$\%$ in ExpRate, $\leq$ 1 error and $\leq$ 2 error, respectively. This shows that our method can improve the performance of GRU-based models and is completely better than the transformer decoder of BTTR.

\subsection{Ablation Study}
We conduct ablation studies to investigate the contributions of different components in the proposed network. The baseline model is a traditional encoder-decoder architecture (DWAP)~\cite{zhang2018deep}, which achieves 50.60$\%$ in ExpRate. 
From Table~\ref{tab:ablation study}, it can be observed that:
\textbf{(1)} The results of “+AAM” show that adding the attention aggregation module to the baseline model further improves the recognition performance of over 2.04$\%$, and the multi-scale attention can benefit the baseline model because of more attentions on small symbols. \textbf{(2)} “+BML” means to equip the baseline model with an additional decoder in inverse decoding direction for mutual learning. From the results, the model can achieve 4.63$\%$ accuracy increment from 50.60$\%$ to 55.23$\%$ after taking advantage of the inverse context information. 
\textbf{(3)}
The results of ABM show that the use of these two modules at the same time generates a cumulative effect, increasing the overall accuracy of the model by a large margin of 6.25$\%$. To this end, we prove that every component in the proposed method can contribute to the overall recognition effectiveness.

\begin{figure}
    \centering
    \includegraphics[width=0.48\textwidth]{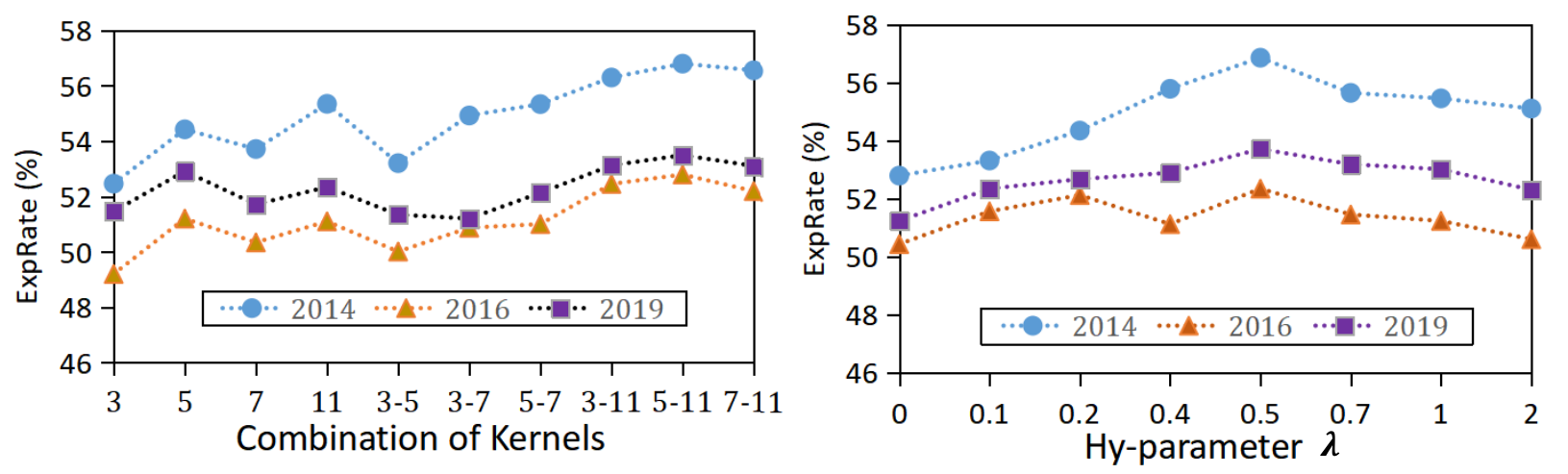}
    \caption{Parameter sensitivity analysis on combination of kernels in AAM and hy-parameter $\lambda$ in Eq. (14). }
    \label{fig:parameter sensitivity}
\end{figure}

\subsection{Parameter Sensitivity}
We perform the sensitivity analysis on the proposed ABM on CROHME 2014 under the univariate setting.
\textbf{Combination of convolution kernels in AAM:} 
Given that convolution kernels of different sizes have distinct receptive fields, we testify the potential values of different convolution kernels in AAM. From Fig.~\ref{fig:parameter sensitivity}(a), the setting of multi-scale kernels can generate better performance on all datasets and the multi-scale attention is beneficial, especially for the symbols with variable scales. Our selection, i.e., combining 5*5 and 11*11, is the most robust strategy. 
\textbf{Hyper-parameter $\lambda$:} $\lambda$ controls the trade-off between the mutual learning loss from two decoding branches and the recognition loss in Eq. (14). In Fig.~\ref{fig:parameter sensitivity}(b), $\lambda$ starts from the small factor (=0.1) to the large (=1), and the performance (ExpRate) increases at the beginning because the other branch brings valuable information from the inverse direction. However, the ExpRate tends to be lower with larger factors (like 0.7) as unstable training may happen. Thus, we set the hyper-parameter $\lambda=0.5$.

\begin{table}[t]
    \centering
    \begin{threeparttable}
    \begin{tabularx}{0.47\textwidth}{l c c c c}
    \hline
        Methods&ExpRate&$\leq$1 error & $\leq$2 error & WER \\
    \hline
     Uni-L2R &50.60& 68.05   & 71.56 & 12.75 \\
     Uni-R2L &49.24 & 68.05   & 71.56 &13.26  \\
     AUM-L2R  & 56.24 & 71.56 & 77.34  & 10.76 \\
     AUM-R2L  &  54.71 &  69.94 & 74.12 & 11.01  \\
     ABM-L2R & \textbf{56.85} & \textbf{73.73} & \textbf{81.24} & \textbf{10.01}\\
     ABM-R2L & 54.86    &  72.01  & 78.90 & 10.86\\
    
    \hline
    \end{tabularx}
    \begin{tablenotes}
     \item[1] “Uni” denotes applying one branch for training.
     \item[2] “AUM” denotes applying uni-directional mutual learning with AAM.
     \item[3] “-L2R” and “-R2L” denotes the results generated from L2R and R2L decoder, respectively.
    \end{tablenotes}
    
    \caption{
    Performance (in $\%$) comparison on different decoding directions. Note that we only use one decoding branch for testing.}
    \label{tab: bidirectionalvssame}
    \end{threeparttable}
\end{table}
\begin{table}[t]
    \centering
    \begin{tabular}{l c c c  c c  }
    \hline
Methods&Prefix-2& Suffix-2 &Prefix-5& Suffix-5   \\
    \hline
    Uni-L2R & 85.29 & 80.02  &72.21 & 67.14  \\ 
    Uni-R2L & 81.22 & 84.47 &67.01 & 71.17  \\
    ABM-L2R & \textbf{88.73} & 84.37 & \textbf{76.35} & 73.10   \\
    ABM-R2L & 83.37 & \textbf{87.12} & 71.60 & \textbf{75.86}  \\ 
    \hline
    \end{tabular}
    \caption{Recognition accuracy (in $\%$) of prefixes-(2, 5) (the first two or five symbols) and suffixes-(2, 5) (the last two or five symbols) on CROHME 2014 test dataset with BML.}
    \label{tab:prefixes and suffixes}
\end{table}
\subsection{Discussion}

\begin{figure*}[th]
    \centering
    \includegraphics[width=\linewidth]{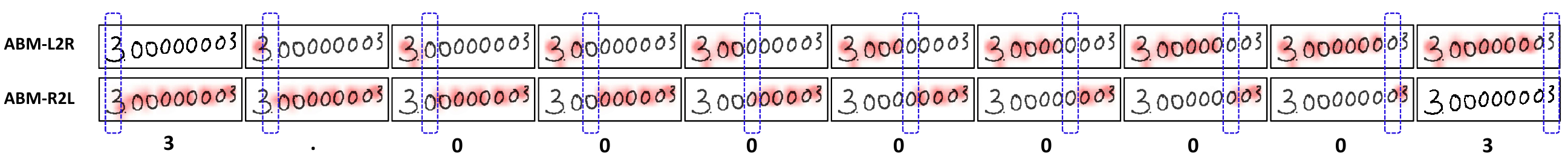}
    \caption{Coverage attention visualization process of translating handwritten mathematical expressions into LaTeX sequences in two directions (L2R and R2L). The blue box indicates the character being decoded in current time step.}
    \label{fig:attention visulation}
\end{figure*}

\subsubsection{Attention Visualization}
To further reveal the internal working mechanism of our proposed method, we visualize the attention process of continuous decoding, as shown in Fig.~\ref{fig:attention visulation}.
Attention weights are visualized in red, and dark red denotes a higher weight in the attention map.
For example, when the decoder translates the fifth character "0", viewing from left to right direction, attention can obtain historical alignment information, and from right to left direction, attention can obtain future alignment information, and eventually accurately locate the current attention position.

\subsubsection{Features Visualization}
\begin{figure}
    \centering
    \includegraphics[width=0.5\textwidth]{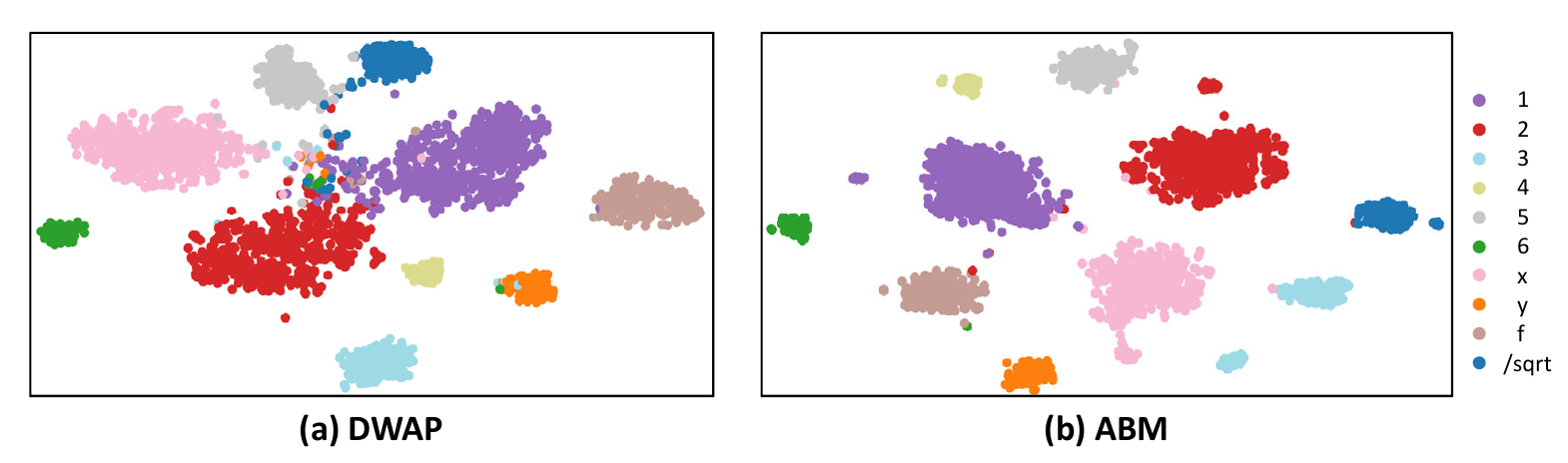}
    \caption{t-SNE Visualisation of DWAP (baseline) and ABM on CROHME 2014.}
    \label{fig:tsne}
\end{figure}

Further, we visualize feature distributions of ten symbols in CROHME 2014 test set by t-SNE~\cite{van2008visualizing}.
We input all previous target symbols to the decoder to predict the current symbols and then visualize the features before the first full connection layer of the classifier.  
Fig.~\ref{fig:tsne} shows that the features of different symbols generated by our method are more separable, compared to the baseline DWAP.


\subsubsection{Evaluating Different Decoding Directions}
We verify the superiority of bi-directional mutual learning over the uni-directional mutual learning and original uni-trained model (Uni-L2R or Uni-R2L). We equip the Uni-directional mutual learning with AAM to form AUM for fair comparison.
From Table~\ref{tab: bidirectionalvssame}, our method (ABM) improves the recognition accuracy by over 6.25$\%$. It turns out that mutual learning between two inverse decoders is more effective compared to uni-direction mutual learning. 

\subsubsection{On Long-distance Dependence}
One major weakness of RNNs is their unbalanced outputs with high-quality prefixes but low-quality suffixes recognition. From Table~\ref{tab:prefixes and suffixes}, there is a margin 5.27 $\%$ between prefix and suffix recognition accuracy in Uni-L2R. After using ABM, the accuracy of the L2R branch increases from 85.29$\%$ to 88.73$\%$ in prefix and from 80.02$\%$ to 84.37$\%$ in suffix. 
Therefore, the L2R branch can learn the decoding knowledge from the R2L branch, and better adapt to long-distance dependence. The recognition abilities in both the directions are improved at the same time.

\begin{table}[t]
    \centering

    \begin{threeparttable}[b]
    \begin{tabularx}{0.48\textwidth}{l p{1.6cm} p{1.3cm} p{1.6cm} }
    \hline
Methods&ExpRate&$\leq$1 error & $\leq$2 error  \\
    \hline
     WAP-MobileNetv2   & 40.61 &60.32 & 67.59  \\
     WAP-MobileNetv2\tnote{$\dagger$} &45.08 & 64.56 & 73.60  \\
     WAP-Xception  & 43.05 & 62.45  & 70.10   \\
     WAP-Xception\tnote{$\dagger$} &46.70 &68.12&75.83    \\
     DWAP-GRU & 50.60 & 68.05 & 71.56  \\
     DWAP-GRU\tnote{$\dagger$}  & \textbf{56.85} & \textbf{73.73} & \textbf{81.24}    \\
     DWAP-LSTM  & 49.64   &   65.62   &   76.06 \\
     DWAP-LSTM\tnote{$\dagger$} & 55.13 &69.95&78.84   \\
     BTTR  &  48.13  &  66.90& 74.30  \\
     BTTR\tnote{*}  &  49.49  &  -& -  \\
     BTTR\tnote{$\ddagger$} &  51.47  &69.23 &76.64  \\
    \hline
    \end{tabularx}
    \begin{tablenotes}
     \item[1] $\dagger$ denotes using AAM and BML modules.
     \item[2] BTTR\tnote{*} denotes applying Bi-trained method without AJS and the results are directly cited from its paper. 
     \item[3] $\ddagger$ denotes only applying BML module as AAM is not suitable for BTTR which adopts parallel decoding.
     \item[4] Results from L2R branch when having two branches.
   \end{tablenotes}
    \end{threeparttable}
    \caption{Performances (in $\%$) of different decoders equipped with our modules.  }
    \label{tab:generality}
\end{table}

\subsubsection{Generality on Different Encoders and Decoders}


We validate the generality of the proposed method on different encoders (MobileNetV2~\cite{sandler2018mobilenetv2}, Xception~\cite{chollet2017xception}, and DenseNet~\cite{huang2017densely}), as well as different decoders (GRU, LSTM, and Transformer). To be fair, all experiments use the same settings. As shown in Table~\ref{tab:generality}, for different encoders, we replace DenseNet with MobileNetV2 and Xception, and their original modles are improved by 4.47$\%$ and 3.65$\%$, respectively, in ExpRate. For different decoders, the GRU, LSTM, and Transformer are improved by 6.25$\%$, 5.49$\%$, and 3.34$\%$, respectively, in ExpRate. We should note that the result of BTTR with one direction training is 48.13$\%$. Therefore, our method is universal on different encoders or decoders.

\subsubsection{Performance on Different Lengths of Expressions}
Further, to explore the ability of our method to decode the sequences of different lengths, we split test datasets into different groups according to the lengths of their corresponding LaTeX sequences and then compare the models for each group. 
Intuitively, the longer the expression, the more difficult the translation.
From Table~\ref{tab:distribution of errors}, BTTR solves the problem of long sequence recognition to a certain extent. This is because it uses bi-directional Transformer as a decoder. But for long formulas with a sequence length greater than 40, its recognition accuracy decreases instead. However, our method can perform well on any sequence length.

\begin{table}[]
    \centering
    \begin{tabular}{l c c c c c}
    \hline
    \multirow{2}*{ Methods}&  \multicolumn{5}{c}{ CROHME 2014 }   \\
    \cline{2-6}
     & [1,10] &[11,20] &[21,30] &[31,40] &[41,$\propto$]  \\
    \hline
     DWAP & 63.91  &54.03   & 47.16 & 43.08 & 26.82    \\ 
     BTTR  &  68.92  &  58.06    &   50.00 & 52.03&20.32    \\
     ABM &   \textbf{72.85} & \textbf{58.46} &\textbf{51.88} & \textbf{53.65} &\textbf{27.66} \\
     
    \hline
    \end{tabular}
      \caption{Accuracy for different lengths of expressions.}
    \label{tab:distribution of errors}
\end{table}

\section{Conclusion}

We propose a novel ABM network for HMER, which uses dual-branch decoders in inverse decoding directions in a mutual learning manner. Experimental results show that the ABM is superior to the state-of-the-arts. Besides, it is applicable to existing decoders including GRU, LSTM and Transformer, and can effectively improve their performances without increasing extra parameters during inference.

\noindent\textbf{Acknowledgment} This work was supported by the Beijing Natural Science Foundation under Grant L191004.

\bibliographystyle{aaai22}
\bibliography{main}

\end{document}